# Shape-Based Single Object Classification Using Ensemble Method Classifiers

Nur Shazwani Kamarudin[*], Mokhairi Makhtar[#], Syadiah Nor Wan Shamsuddin[#], Syed Abdullah Fadzli[#]

[#]*Faculty of Informatics and Computing, Universiti Sultan Zainal Abidin, Tembila, Terengganu, Malaysia*
*E-mail: wanikamarudin@gmail.com, mokhairi@unisza.edu.my, syadiah@unisza.edu.my, fadzlihasan@unisza.edu.my*

*Abstract*— Nowadays, more and more images are available. Annotation and retrieval of the images pose classification problems, where each class is defined as the group of database images labelled with a common semantic label. Various systems have been proposed for content-based retrieval, as well as for image classification and indexing. In this paper, a hierarchical classification framework has been proposed for bridging the semantic gap effectively and achieving multi-category image classification. A well-known pre-processing and post-processing method was used and applied to three problems; image segmentation, object identification and image classification. The method was applied to classify single object images from Amazon and Google datasets. The classification was tested for four different classifiers; BayesNetwork (BN), Random Forest (RF), Bagging and Vote. The estimated classification accuracies ranged from 20% to 99% (using 10-fold cross validation). The Bagging classifier presents the best performance, followed by the Random Forest classifier.

## I. INTRODUCTION

Image classification has been one of the most extensively studied fields in the pattern recognition community. Many factors can affect the complex processes in image classification [1]. Image classification is the process of labeling images into one of a number of predefined categories [2].

The ensemble method is popular among machine learning research field because its algorithm has the capability of combining a set of individual classifiers (called base learners). New data points will be produced by taking a weighted or unweighted vote of the predictions and provide a better result. Normally the ensemble method will improve the prediction performance. The main idea behind the ensemble methodology is to weigh several individual classifiers and to combine them in order to obtain a classifier that outperforms every one of them [15].

The tuning process in finding optimum model parameters is important. Each model generated must be trained to find the most relevant attributes and model parameters in producing a quality model. The tuning process involves selecting optimum model parameters such as number of folds for cross validation and type of classifier. Selection of the optimum attributes from the data set is also another step of the tuning process. This will be repeated until the right combination of parameters is selected to generate the best model [16].

The experiment was performed on the Amazon and Google images which consisted of 11 features for each single image. This experiment used Weka with 10-fold cross validation to run the classification experiment and the classifiers chosen were:
  a) weka.classifiers.trees.RandomForest,
  b) weka.classifiers.meta.Vote,
  c) weka.classifiers.bayes.BayesNet.
  d) weka.classifiers.meta.Bagging

From the listed classifiers above, Random Forest and Bagging represent the ensemble classifiers. Random Forest is based on the combination of tree models, which is quite sensitive to variations in the training data, while Bagging is a method that creates diverse models on different random samples [17].

Meanwhile, BayesNet uses Bayes' rule that represents knowledge about an uncertain domain. A BayesNet reflects a simple conditional independence statement. Thus, each variable is independent of its non descendents in the graph, given the state of its parents. Furthermore, Multi Class Classifier is a classification task with more than two classes and Vote is a binary classifier that works by (one-versus-one-voting) [18].

This paper elaborates classifier algorithms used in the experiments. As mentioned earlier, those classifiers are BayesNetwork (BN), Random Forest (RF), Bagging and Vote. All these classifiers classify the selected dataset and the results are then compared. Fig. 1 shows the image pre-processing steps involved in this experiment.

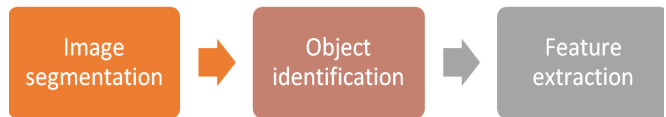

Fig. 1 Image pre-processing

The crucial components before the classification process are image segmentation, object identification and feature extraction. The important goal of segmentation is to distinguish semantically significant parts of a picture and classify the pixels that having a place with such segments [3]. This paper used one of the global thresholding methods, which is known as Otsu method.

Object identification is another step prior to image classification. With the built-in Image Processing Toolbox in MATLAB, image processing becomes much easier. One of the provided functions in the toolbox for object identification is 'bwlabel', which is capable of detecting connected components in 2D binary image [4].

The component extraction process is the place the rich substance of pictures changes into different substance highlights. In [5] characterized that component extraction is the way toward creating elements to be utilized as a part of the determination and grouping errands.

## II. MATERIAL AND METHOD

The process begins with pre-processing or many other authors' state as the image acquisition process and is followed by the segmentation process. Pre-processing must be done in order to remove noise and enhance the image quality. Meanwhile, segmentation is the process to remove background from the region of interest (ROI) in an image. Feature extraction is the calculation of image features after the segmentation process is done. Feature selection sometimes gives an issue to the researcher in order to choose the best set of features. Then, the classification process will classify all the selected features.

This paper follow classification process consists of the following steps.

### A. Pre-Processing

One of the classical examples of multichannel information processing is colour image pre-processing and segmentation. Ever since colour intensity information is generally manifested in the form of admixtures of dissimilar colour components, the task of colour image processing includes a vast amount of processing overhead. Furthermore, the relative scopes of the component colours and their inter-correlations also exhibit nonlinear features. The main target of image pre-processing is to enhance the quality of the input image such as noise removal, image masking, main component analysis, to locate the data of interest, atmospheric correction, noise removal and image transformation [19].

### B. Detection and Extraction of an Object

Detection includes the detection of position and other characteristics of the moving object image obtained from camera and in the extraction, the detected object estimated the trajectory of the object in the image plane. For feature analysis process, it started with the feature extraction and finished with feature classification (image classification based on the image feature). The main function of the module is to extract a representative set of features of the images. The aim of this step is to replace the high-dimensional images with lower-dimensional features that capture the main properties of the images, and to enable the model to work on the data with limited memory and computational resources. The system loads the pre-computed image features from the text files stored and the next step of processing is the feature database [20].

### C. Training

Selection of the particular attribute which best describes the pattern.

### D. Classification of the Object

Object classification step categorizes detected objects into predefined classes by using suitable method that compares the image patterns with the target patterns. For the comparison and evaluation of the classification method, appropriate datasets were required. The data repository should consist of enough images in each category. The dataset consisted of images downloaded from the Internet.

### E. Amazon Images

Following are the steps involved in the process of collecting images from Amazon.com. The python code developed was run and the process of downloading images started. At the beginning of the process, a connection to the Amazon website was made. After that, the python opened a '.csv' file to save all the data retrieved in that file. Then, the python code selected a category from the website. If it matched well with the 'URL' in the code, all data were extracted and saved in the '.csv' file. Otherwise, the python would re-select the category. The saved data finally could be used and the researcher would continue the downloading process for the next categories. The average time needed to download 1000 images was about 30 minutes. The researcher took about two days to complete the downloading process. Then, the 37 images were divided into training and testing groups. In the training process, the images went through three major steps: a) image segmentation, b) feature extraction, and c) image classification in order to obtain the image data. This process aimed to obtain the data of the image and classified them based on their features.

With the primary objective of this research, the Amazon product was selected to be the training and testing image. An amazing product with a standard image such as white background, made it easier to work with later. The Amazon dataset was created manually using the python script. Python was chosen because it was the simplest and the easiest programming language. It is also widely used because of its algorithm, which is uncomplicated. A python script was developed in order to generate data from Amazon.com. The script was able to download images directly from the Amazon.com web page.

*F. Google Images*

Google images are another dataset used in this research. It was downloaded by the research team manually from the Google image. The objective of using this dataset was to compare the results obtained from Amazon and Google datasets as both datasets presented pictures in different ways. The Google images used in the research were selected based on the background and the position of the image in the picture. The images selected did not use white background, as found in the Amazon images, which had been the main focus in the image that must be the objects of beg, shoes, dress, etc. The following screenshots shows the images from Google dataset.

*G. MATLAB*

MATLAB ("MATrix LABoratory") is a tool for numerical computation and visualization. MATLAB® is a high-level language and an interactive environment for numerical computation, visualization and programming. It has the ability to analyse data, develop algorithms, create models and applications. The MATLAB language is dedicated to matrix calculations, and has been optimized in this perspective. The variables are handled as the priority is real or complex matrices.

In this study, MATLAB provided an image processing toolbox with many powerful and very efficient image-processing functions. This research focused on segmenting, labeling and extracting data from images using numbers of functions available from MATLAB [13], [14].

*H. Weka*

Weka is open source software under the GNU General Public License (see Fig. 2). It was developed at the University of Waikato, New Zealand. "WEKA" stands for Waikato Environment for Knowledge Analysis. It is written using object oriented language Java. Weka provides both implementation of state-of-the-art data mining and machine learning algorithms. It contained modules for data pre-processing, classification, clustering, and association rule extraction. Accuracy provided by each tool was compared in order to determine the best tool and technique for classification.

The main features of Weka include:
- Data pre-processing tools
- Classification/regression algorithms
- Clustering algorithms
- Attribute/subset evaluators + 10 search algorithms for feature selection.
- Algorithms for association rules
- Graphical user interfaces
- The explorer" (exploratory data analysis)
- The Experimenter" (experimental environment)
- The Knowledge Flow" (new process model inspired interface)

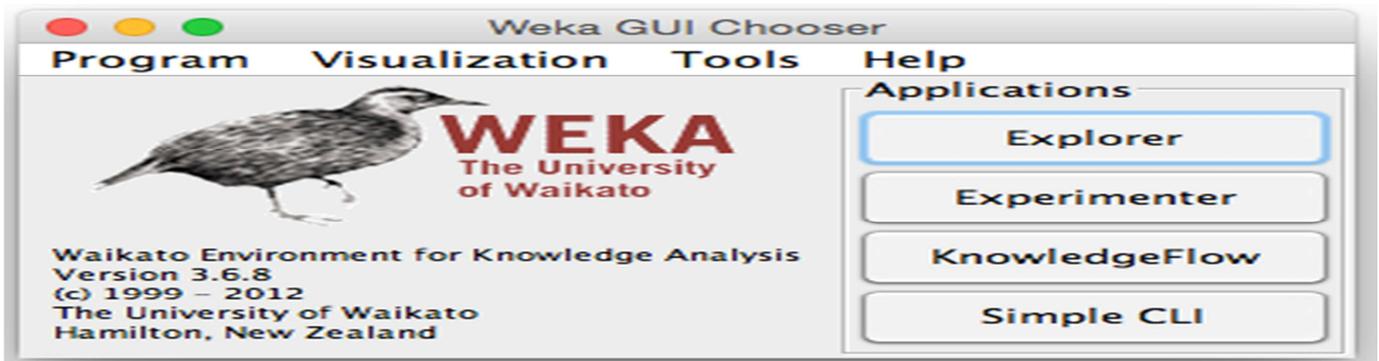

Fig. 2 Weka tools

Classification is one of the data mining tasks that learn from a collection of cases in order to accurately predict the target class for new cases. To perform classification, some machine learning techniques can be used. In order to perform the classification through different techniques that will be discussed in this section.

There are two main phases in a classification system: training and testing. Training is the cognitive operation of defining criteria by which characteristics are distinguished. In this process, the classifier learns its own classification rules from a training set. In the training process, images are captured and stored in a database.

*I. Bayesnetwork (Bn)*

This is the outline of the general options used by BayesNetwork classifier [6]:
1. Debug-If set to true, the classifier may produce additional output info to the console.
2. BIFFile-Set the name of a file in BIF XML format. A Bayes network learned from data can be compared with the Bayes network represented by the BIF file. Statistics calculated are o.a. the number of missing and extra arcs.
3. SearchAlgorithm-Select method used for searching network structures.
4. UseADTree-When ADTree (the data structure for increasing speed on counts, not to be confused with the classifier under the same name) is used, learning time typically goes down. However, because ADTrees are memory intensive, memory problems may occur. Switching this option off makes the structure learns algorithms slower and runs with less memory. By default, ADTrees are used.
5. Estimator-Select Estimator algorithm for finding the conditional probability tables of the Bayes Network.

*J. Random Forest*

Random Forest is a multi-way classifier with the existence number of trees, where the trees are grown using some sort of randomization. It is based on the joint induction of shape features and tree classifiers [7], [11], [12]. The leaf nodes of each tree are labeled by the estimation of the posterior distribution over the image classes. Each single internal node consists of a test that differentiates the space of the data to be classified. The classification process continues by sending the image down to every tree until it reaches the leaf. The randomness points can be inserted at specifically two main points during the training process, which is the sub-sampling of the training data and selecting the node test [8]. The basic algorithm of Random Forest is shown in Fig. 3. The randomness points can be inserted at specifically two main points during the training process, which is the sub-sampling of the training data and selecting the node test.

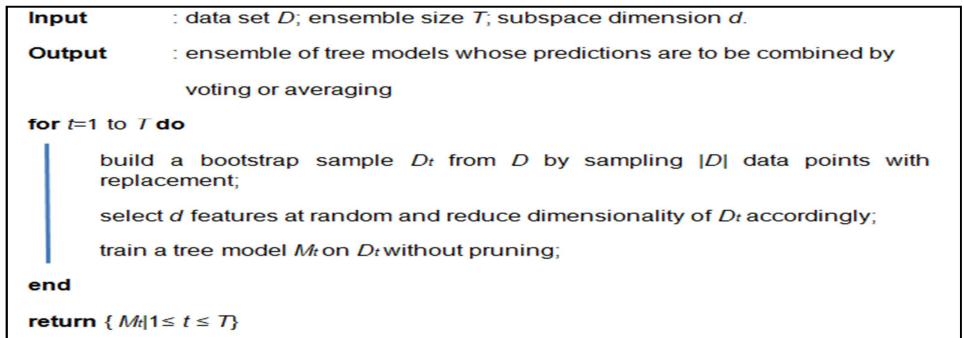

Fig. 3 Random Forest algorithm

*K. Bagging*

Bagging is a short form of 'bootstrap aggregating', which is simple but highly effective ensemble method that creates diverse models on different random samples of the original dataset. These samples are taken uniformly with replacement and are known as bootstrap samples. The algorithm in Fig. 4 gives the basic Bagging algorithm, which returns the ensemble as a set of models [9].

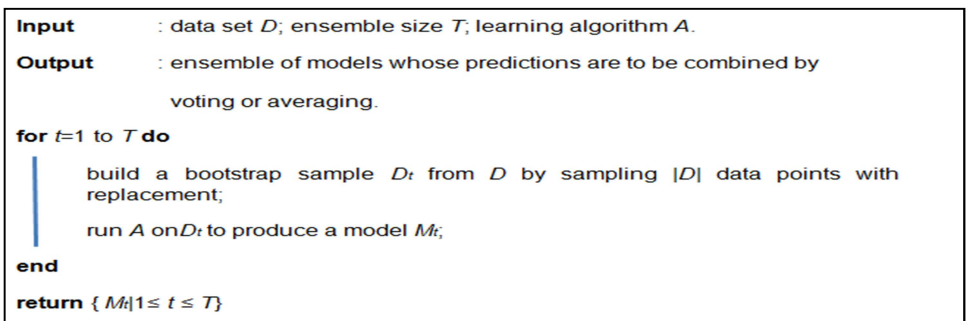

Fig. 4 Bagging algorithms

*L. Vote*

This is the outline of the general options used by Vote classifier [6]:
1. Debug-If sets to true, the classifier may produce additional output info to the console.
2. Seed-The random number of seed to be used.
3. CombinationRule-The combination rule to be used.
4. Classifiers-The base classifiers to be used.
5. PreBuiltClassifiers-The pre-built serialized classifiers to be included. Multiple serialized classifiers can be included alongside those that are built from scratch when this classifier runs. Note that it does not make sense to include pre-built classifiers in a cross-validation, since they are static and their models do not change from fold to fold.

*M. Dataset*

The datasets used in this experiment were collected from Amazon and Google images. The Amazon dataset was created manually using Phyton script and Google dataset was collected manually from the Google search engine [10].

Fig. 5 and Fig. 6 show the samples of Amazon and Google images after subjected to the segmentation process. The Otsu method works by selecting a threshold automatically from a grey level histogram. Even though the method is simple and easy, Otsu method can still give better results which depending on the nature of the image [10].

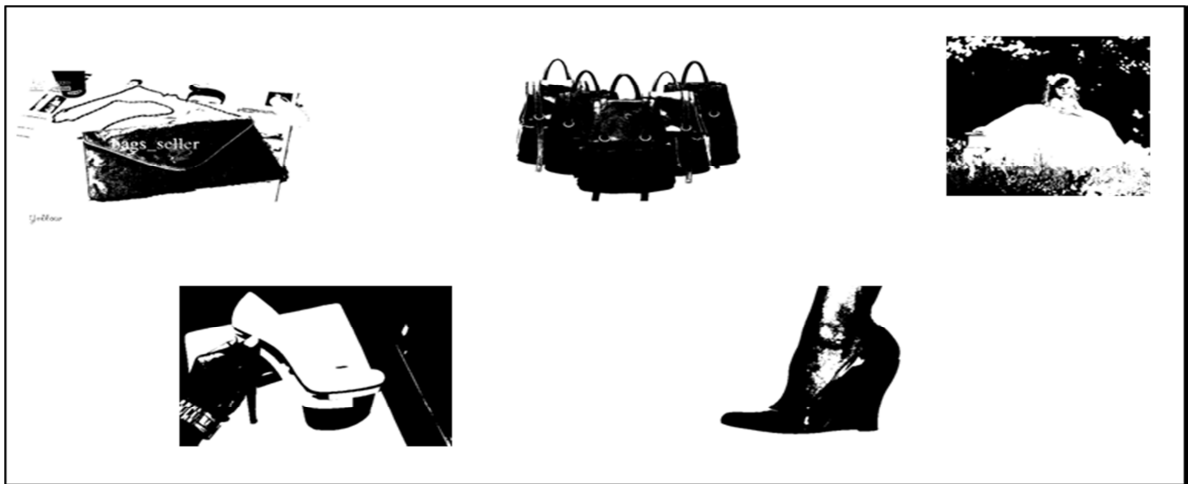
Fig. 5  Google dataset sample

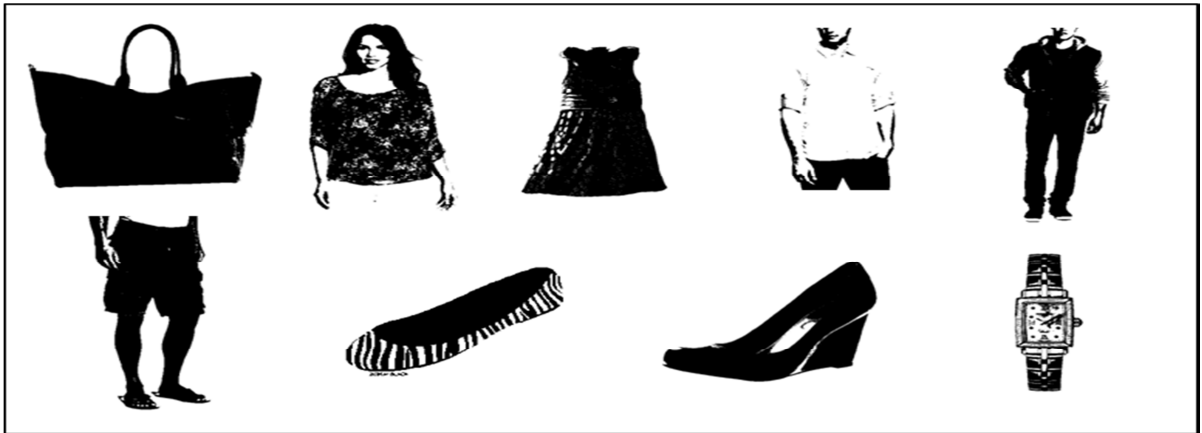
Fig. 6  Amazon dataset sample

After that, all images were subjected to another process, which was object identification using 'bwlabel' built-in function in MATLAB toolbox. The algorithms involve 4 steps, starting with run-length that encodes the input image. Then, the algorithm scans the runs by assigning preliminary labels and recording label equivalences in a local equivalence table. After that, it resolves the equivalence classes and finally re-labels the runs based on the resolved equivalence classes.

For feature extraction process, it was done with MATLAB procedure using the regionprops function from the Image Processing toolbox to extract 11 image features such as area, major axis length, minor axis length, eccentricity, orientation, convex area, filled area, Euler number, EquivDiameter, solidity and extent.

### III. RESULTS AND DISCUSSION

From the listed classifiers in Section 2, Random Forest and Bagging represent the ensemble classifiers. Random Forest is based on the combination of tree models, which are quite sensitive to variations in the training data, whereas Bagging is a method that creates diverse models on different random samples.

Table I and Fig. 7 show the result for single object images that underwent classification process using four different classifiers. For both dataset, Bagging classifier gave the highest accuracy compared to the others with 99.67% on Amazon and 99.18% on Google. It is followed by Random Forest with the percentage of accuracy of 98.45% on Amazon. It differed for Google as BayesNetwork gave higher accuracy compared to Random Forest with 97.96% over 92.65% respectively. For Vote classifier, both datasets gave the lowest result.

TABLE I
MODEL ACCURACIES FOR SINGLE OBJECT IMAGE

|  | Random Forest [C2] | Vote [C3] | BayesNetwork [C4] | Bagging [C5] |
| --- | --- | --- | --- | --- |
| Amazon dataset | 98.45 | 20.70 | 80.33 | 99.67 |
| Google dataset | 92.65 | 20.41 | 87.96 | 99.18 |

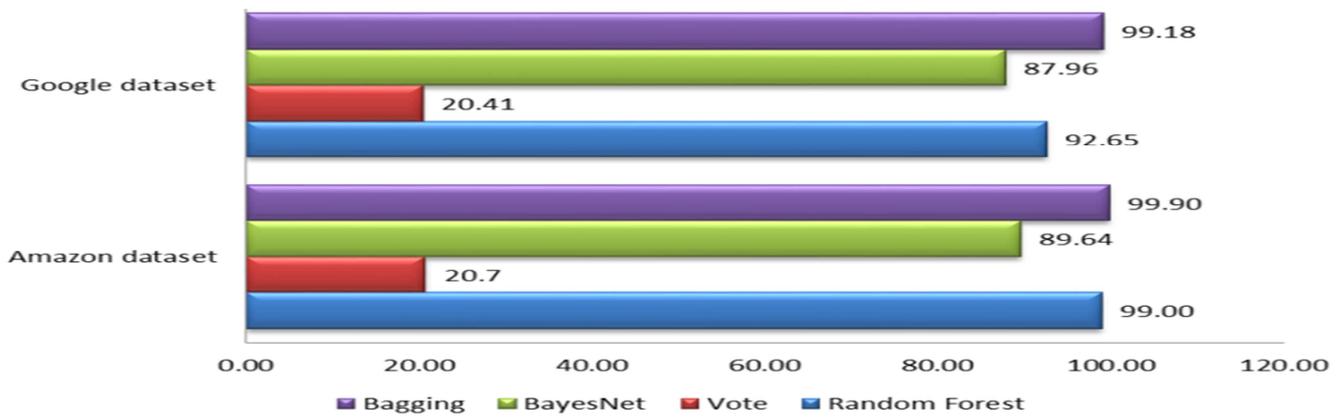

Fig. 7 Accuracy graph for single object image dataset

From the results above, it can be concluded that the ensemble methods affected the classification accuracy. An ensemble is largely characterized by the diversity generation mechanism and the choice of its combination procedure.

IV. CONCLUSION

In this paper, classification techniques for the single object image of multiple categories have been reviewed. Emphasis has been given to those techniques in ensemble method. The results provided proved that the ensemble method gives higher classification accuracy compared to other methods. It shows that the method works really well with the single object images from Amazon and Google datasets. The classifier is able to achieve nearly 99% accuracy. In conclusion, we remain firm that our technique demonstrated usefulness and effectiveness for pre-processing and classification function, particularly by considering its simplicity.

ACKNOWLEDGMENT

This work is partially supported by UniSZA (Grant Code. UniSZA/13/GU(29)).